\documentclass[journal]{IEEEtran}
\usepackage{graphicx}
\usepackage{flushend}
\ifCLASSINFOpdf
\else
\fi
\newcommand{\datasetname}{NASTaR}
\begin{document}
%
\title{\datasetname{}: A NovaSAR-Based Automated Ship Target Recognition Dataset}
%
%

\author{Benyamin~Hosseiny,~\IEEEmembership{Member,~IEEE,}
        Kamirul~Kamirul,
        Odysseas~Pappas,
        and~Alin~Achim,~\IEEEmembership{Senior~Member,~IEEE}
\thanks{The authors  are with the VI-Labs, University of
Bristol, BS1 5DD Bristol, U.K.  (e-mail: ben.hosseiny@bristol.ac.uk; kamirul.kamirul@bristol.ac.uk;
o.pappas@bristol.ac.uk; alin.achim@bristol.ac.uk)}
\thanks{ This work was supported in part by the Engineering and Physical Sciences Research Council under Grant EP/X525674/1 EPSRC Impact Acceleration Account - University of Bristol 2022 and in part by the Dstl under the DASA Defence Rapid Impact Open Call. The work of Kamirul Kamirul was supported by Lembaga Pengelola Dana Pendidikan (LPDP), Ministry of Finance of the Republic of Indonesia. }}

%
%

\markboth{IEEE Geoscience and Remote Sensing Letters, January~2026}%
{Shell \MakeLowercase{\textit{et al.}}: Bare Demo of IEEEtran.cls for Journals}
%



\maketitle

\begin{abstract}
We introduce the \underline{N}ovaSAR \underline{A}utomated \underline{S}hip \underline{Ta}rget \underline{R}ecognition (\datasetname{}) dataset. This dataset comprises of 3415 ship patches extracted from NovaSAR S-band imagery, with labels matched to AIS data. It includes distinctive features such as 23 unique classes, inshore/offshore separation, and an auxiliary wake dataset for patches where ship wakes are visible. We validated the dataset’s applicability across prominent ship-type classification scenarios using benchmark deep learning models. Results demonstrate over 60\% accuracy for classifying four major ship types, over 70\% for a three-class scenario, more than 75\% for distinguishing cargo from tanker ships, and over 87\% for identifying fishing vessels. The \datasetname{} dataset is available at  https://doi.org/10.5523/bris.2tfa6x37oerz2lyiw6hp47058, while relevant codes for benchmarking and analysis are available at https://github.com/benyaminhosseiny/nastar.
\end{abstract}


%
\IEEEpeerreviewmaketitle

\section{Introduction}
%
%
%
%

\IEEEPARstart{S}{ynthetic} Aperture Radar (SAR) operate across various frequency bands, such as X, C, L, and more recently, the S-band, each offering distinct trade-offs. Higher-frequency bands (e.g., X and C bands) provide finer spatial resolution, enabling detailed object delineation but suffer from limited penetration and lower SNR in cluttered environments. Lower-frequency bands (e.g., the L-band) offer deeper penetration and broader swath coverage, but at the cost of reduced spatial resolution. S-band SAR, exemplified by platforms like NovaSAR and NiSAR, offers a compromise between resolution and coverage. Its moderate wavelength allows for improved SNR and wider swath acquisition compared to the X- and C bands, while retaining sufficient resolution for object-level analysis \cite{Kamirul2024}. These characteristics make S-band SAR particularly promising for large-scale maritime surveillance, where both spatial detail and temporal coverage are critical.



SAR’s coherent imaging and side‑looking geometry introduce speckle and geometric distortions that complicate interpretation, while maritime backscatter varies with ship shape and sea state. These factors make automatic target recognition (ATR) challenging, and traditional model‑ or feature‑based methods often generalize poorly \cite{awais2025survey}.
Deep learning (DL) offers strong performance by learning features directly from data, enabling robust detection and classification. Despite this, DL methods require large, well‑annotated datasets, which remain difficult to produce due to interpretation complexity and radar signature variability.

Benchmark datasets play a critical role in advancing DL research by providing standardized evaluation protocols and facilitating reproducibility. While several SAR datasets exist for ship detection and classification, most are based on X- or C-band imagery and focus on binary detection or coarse classification tasks \cite{zhang2024development}.
There is a notable gap in publicly available datasets tailored to S-band SAR imagery, especially for fine-grained ship-type classification. Given the unique imaging characteristics of the S-band and its growing adoption in maritime surveillance, a dedicated benchmark is essential to support model development, cross-band generalization studies, and operational deployment.

In this paper, we introduce \datasetname{} (NovaSAR Automated
Ship Target Recognition), a benchmark dataset for ship-type classification in S-band SAR images acquired from the NovaSAR satellite. The dataset includes annotated samples across multiple ship categories, enabling tasks that include (i) ship detection and classification: distinguishing between fishing, cargo, tanker, and other vessel types; (ii) wake identification: recognizing ship wakes and their patterns for ship classification and motion inference; (iii) model benchmarking: evaluation of the performance of various DL architectures for S-band SAR imagery.


The remainder of this paper is organized as follows. Section~\ref{section:Related Works} reviews public ship recognition datasets from SAR imagery and related state-of-the-art analysis techniques. Section~\ref{section:Dataset} describes the construction of \datasetname{}, its key features, and statistical insights, including class distributions and variations in object shapes and orientations. Section~\ref{section:Benchmark Experiments} presents benchmark results, evaluating deep learning models on the \datasetname{} dataset, with concluding remarks in Section~\ref{section:Conclusion}.

\section{Related Works}\label{section:Related Works}
\subsection{ Existing SAR Ship Recognition Datasets}

Publicly available SAR datasets for ship ATR remain limited. Early resources include the HR‑SAR dataset \cite{xing2013ship}, containing 450 high‑resolution TerraSAR‑X samples across three ship classes. Another widely used dataset is OpenSARShip \cite{huang2017opensarship, li2017opensarship2}, derived from Sentinel‑1 SLC and GRD data, though it suffers from severe class imbalance. xView3‑SAR \cite{paolo2022xview3} focuses on detecting dark vessels in Sentinel‑1 imagery, while high‑resolution Gaofen‑based datasets such as FUSAR \cite{hou2020fusar} and MTCD \cite{ma2018ship} provide detailed ship signatures. However, none of these datasets include S‑band SAR imagery.

Datasets dedicated to ship wakes are even more limited. Existing resources primarily address wake detection rather than classification. OpenSARWake \cite{xu2024opensarwake} supports wake detection using Sentinel‑1, Gaofen, and ALOS‑PALSAR data, while SynthWakeSAR \cite{rizaev2022synthwakesar} provides simulated wake imagery generated from hydrodynamic models. To the best of our knowledge, no publicly available dataset focuses on ship identification using real S‑band wake patterns.

\subsection{Deep learning Techniques in Ship Target Recognition}



Recent research efforts have been undertaken to improve SAR ship ATR by addressing main challenges, such as class imbalance, noisy, varied shapes, and sparse backscattering. For example, to mitigate the class imbalance problem in the FUSAR and OpenSARShip datasets, Zhang et al., \cite{zhang2021imbalanced} proposed a lightweight customized CNN model and a training process based on identifying the center of each class in the deep feature space, followed by gradually balanced sampling. To address the large shape and size variations among different ship types and the significant inter-class overlap, a multiscale feature attention mechanism was introduced in \cite{wang2023sar}. This approach improved classification accuracy through an adaptive weighting technique that effectively emphasizes features from different scales. To enhance model robustness and reliability in ship recognition, Zheng et al., \cite{zheng2022metaboost} developed a MetaBoost-based ensemble learning model, which demonstrated improved performance across diverse SAR scenes.

\section{The \datasetname{} Dataset}\label{section:Dataset}
\begin{figure}[!t]
    \centering
    \includegraphics[width=0.8\linewidth]{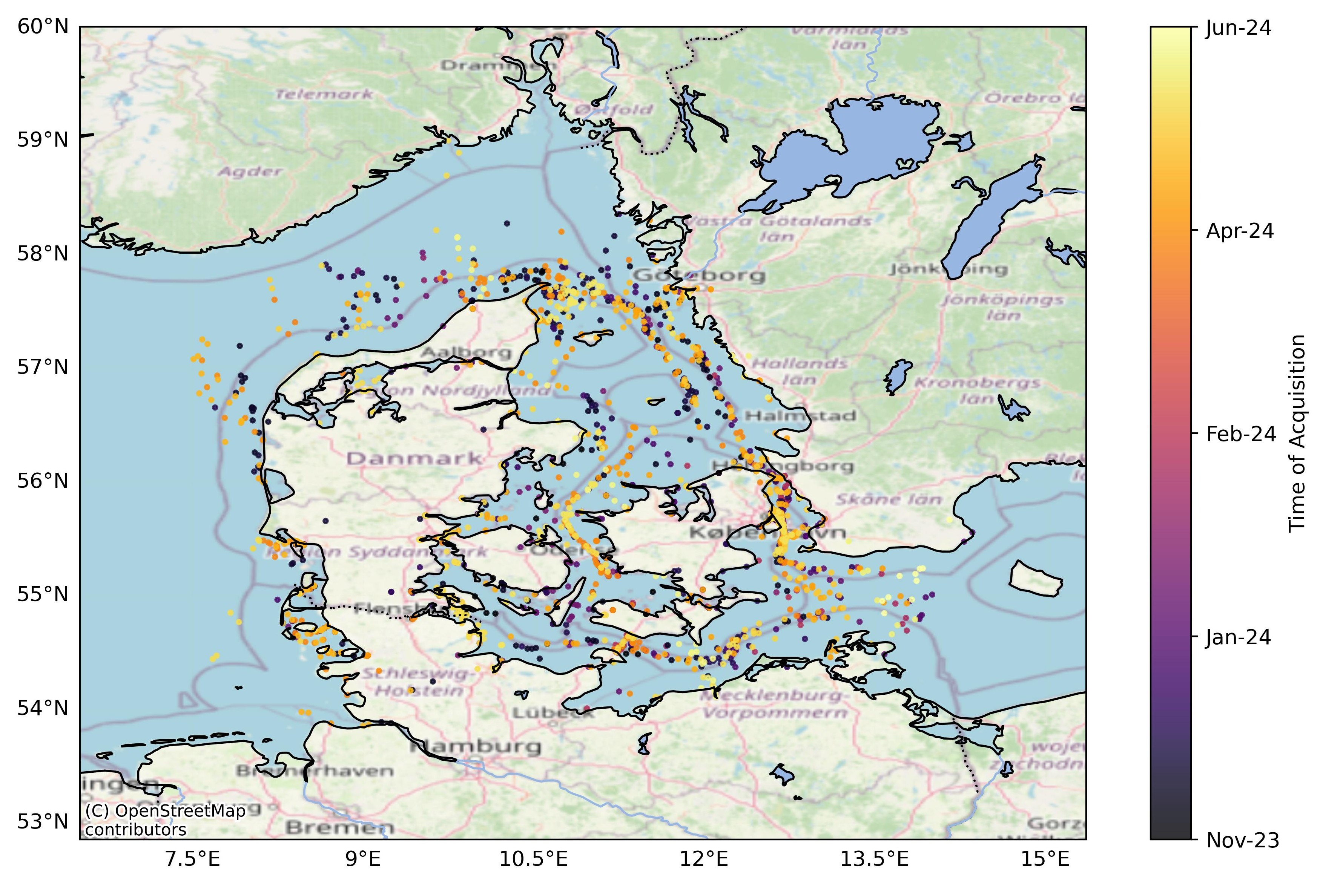}
    \caption{Geographic and temporal distribution of the constructed dataset}
    \label{fig:geomap}
\end{figure}

NovaSAR was launched in 2018 and aims to demonstrate the potential of low-cost, miniature SAR satellites for maritime monitoring and land use applications. It is an S-band (3.2 GHz) satellite producing medium resolution (6-30m) images. To construct \datasetname{}, 624 NovaSAR images captured over the North sea and the Danish straits between November 2023 and June 2024 were used. The images were acquired in Stripmap mode with 6m resolution and are ground-range detected (GRD). The temporal and spatial distribution of ship presence during this period is illustrated in Fig.~\ref{fig:geomap}.

To generate ship‑type ground truth, ships in the NovaSAR images were matched with AIS data from the Danish Maritime Authority \cite{DMA_AIS_Data}, whose ground‑based receivers provide accurate coverage in the study area. AIS samples were filtered to the same geographic region and to a 100‑second window around each SAR acquisition. 
We also assessed AIS–SAR matching uncertainty due to reporting delays, vessel motion, and geolocation error; as shown in Supplementary Material S4, the combined uncertainty remains well within the patch size, and ambiguous multi‑vessel cases were removed. Alongside position and timestamp, the AIS data provide attributes such as speed over ground (SOG), heading, vessel type, and dimensions.


Ship patches were generated by mapping AIS positions to the corresponding image coordinates, producing 512×512‑pixel (1280×1280 m²) crops. An initial 3415 patches were extracted. To separate inshore and offshore samples, the distance to the nearest landmass was computed using Natural Earth maps \cite{Natural_Earth_vec}; applying a 500‑m threshold yielded 1960 inshore and 1455 offshore patches. The computed distances are included to support user‑defined splits.

\begin{figure}[!t]
    \centering
    \includegraphics[width=0.8\linewidth]{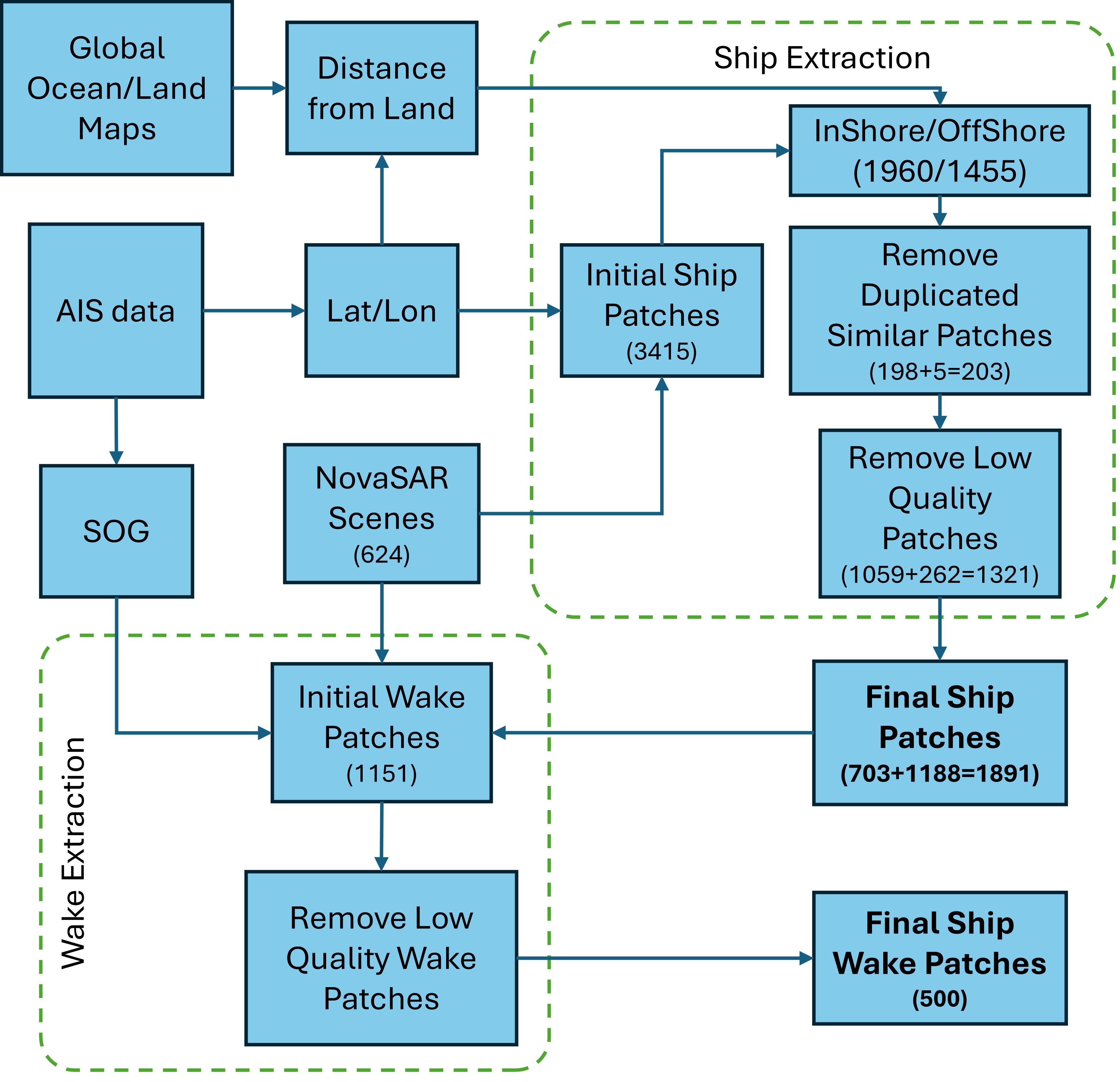}
    \caption{Block diagram of the semi-automated patch extraction pipeline.}
    \label{fig:flowchart}
\end{figure}

Two data filtering steps were applied: one to remove duplicate patches where multiple ships were located close together, and another to eliminate low-quality patches with excessive noise or unclear ship patterns. As a result, 198 inshore and 5 offshore duplicate patches were removed, along with 1059 inshore and 262 offshore low-quality patches. 
A detailed breakdown of the filtering steps and the resulting per-class sample counts is provided in Supplementary Material~S1 and S2. The largest reduction occurs for fishing vessels, primarily due to their smaller size and weaker or partially obscured signatures in S-band SAR imagery. Despite this reduction, the relative class frequency ordering remains consistent, indicating that the filtering process does not introduce substantial class imbalance or bias.
The final dataset consists of 1891 ship patches, including 703 inshore and 1188 offshore samples.
To extract ship wake samples, AIS-derived SOG data was used to identify ships moving faster than 1 m/s, which are likely to generate visible wake patterns. Initially, 1151 wake candidate patches were extracted. After filtering out low-quality samples (where wake visibility was poor due to a number of factors including sea state and SAR viewing geometry), 500 wake patches were retained. Each wake patch is sized at 1024×1024 pixels (2560×2560 m² on the ground), with the ship positioned in a corner rather than the center to better capture the extended wake pattern. The flowchart of semi-automated extraction and labeling of ship and ship wake patches is shown in Fig.~\ref{fig:flowchart}.

\begin{figure}[!t]
    \centering
    \includegraphics[width=0.95\linewidth]{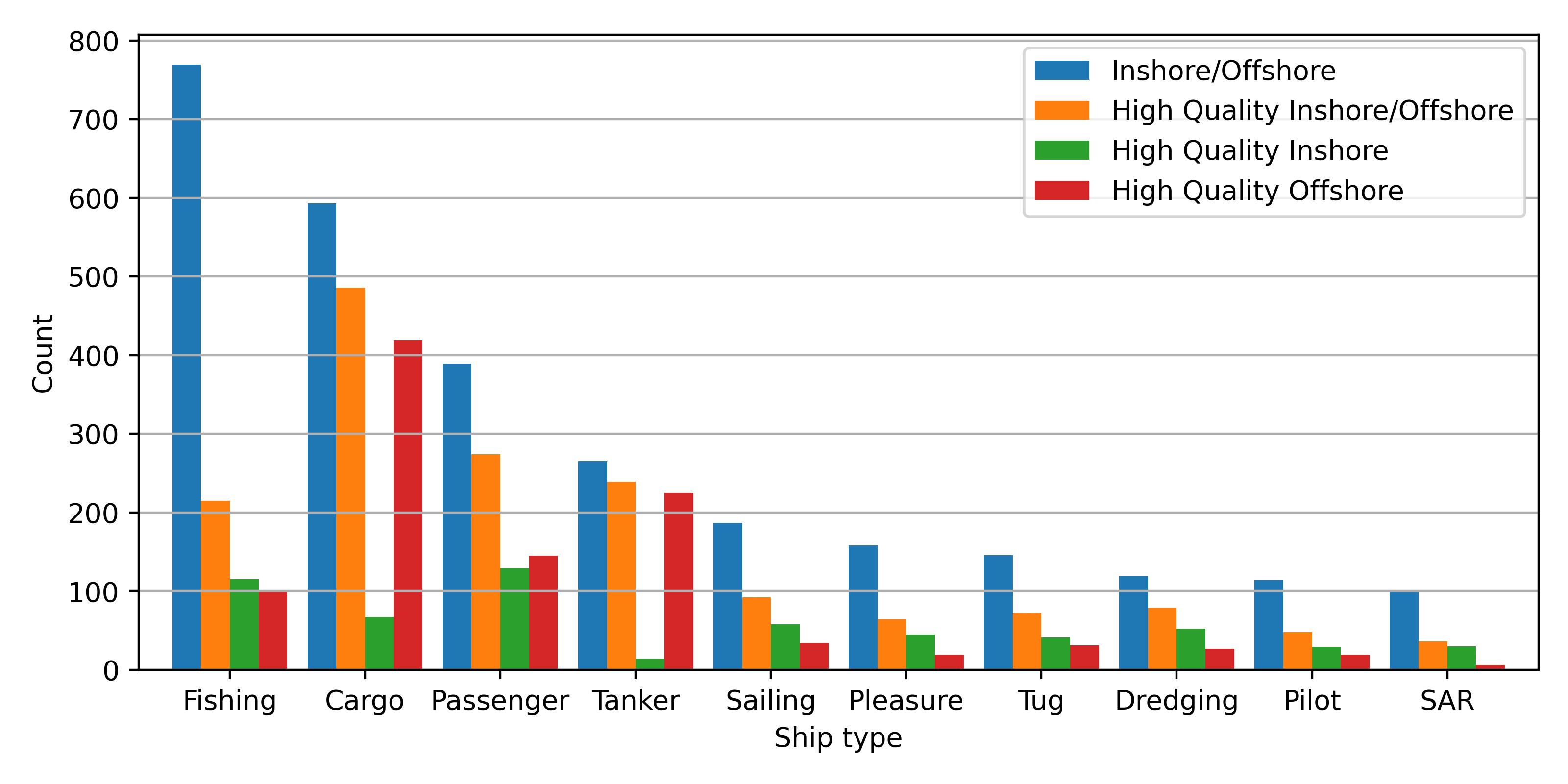}
    \caption{Class distribution of the extracted ship images.}
    \label{fig:shipbarchart}
\end{figure}

\begin{figure}[!t]
    \centering
    \includegraphics[width=0.93\linewidth]{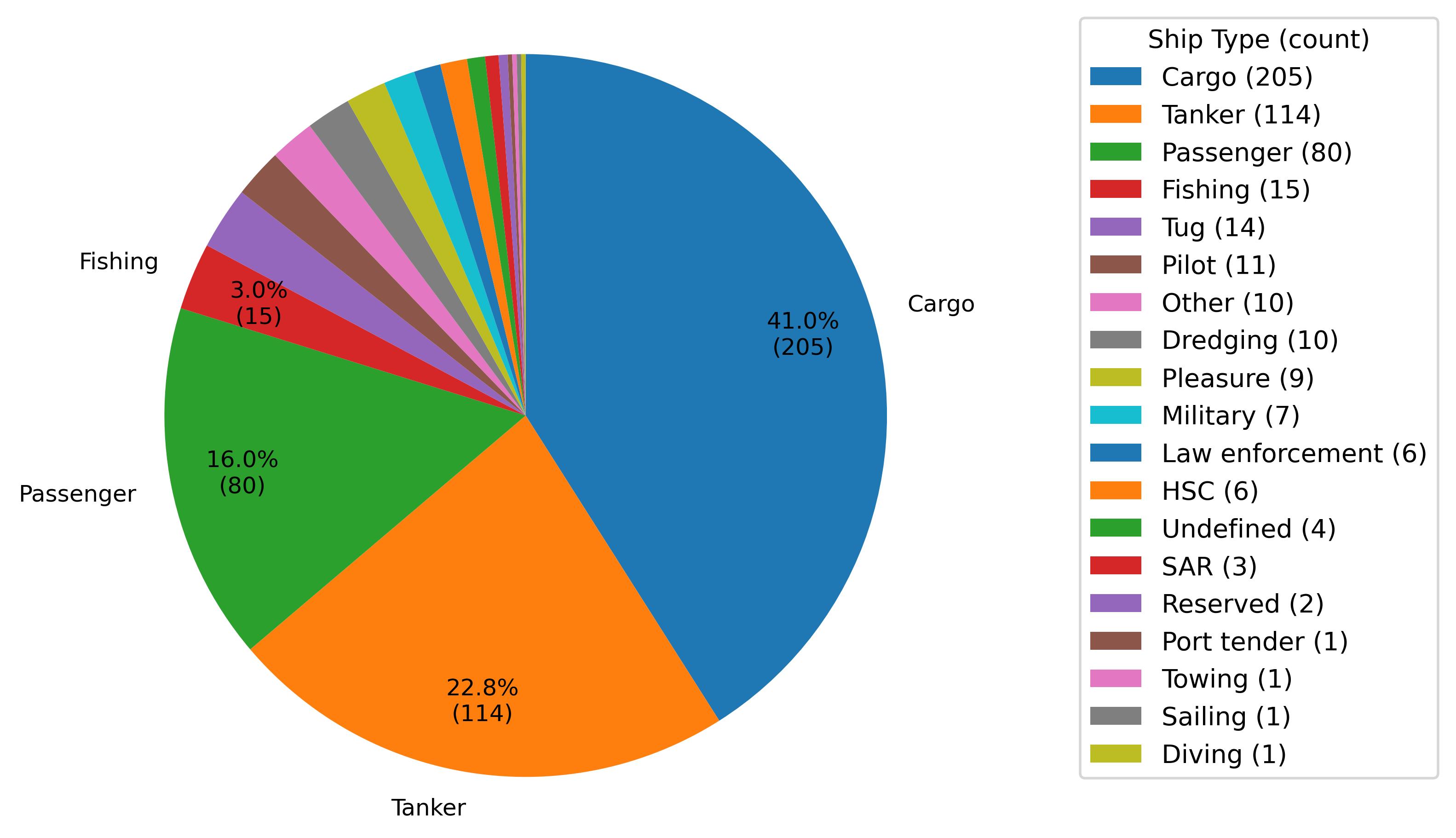}
    \caption{Corresponding class distribution of extracted ship wakes.}
    \label{fig:wakepiechart}
\end{figure}

The complete extracted dataset comprises over 23 distinct ship types, including cargo, tanker, fishing, passenger, and sailing vessels.
Fig.~\ref{fig:shipbarchart} shows the distribution of the top 10 ship types in the dataset. Fishing vessels are the most prevalent in the initial dataset, while cargo ships dominate after filtering, particularly in offshore regions. Cargo ships also constitute the majority of samples in the ship wake dataset, followed by tanker and passenger ships (Fig.~\ref{fig:wakepiechart}).

\begin{figure}[!t]
    \centering
    \includegraphics[width=0.93\linewidth]{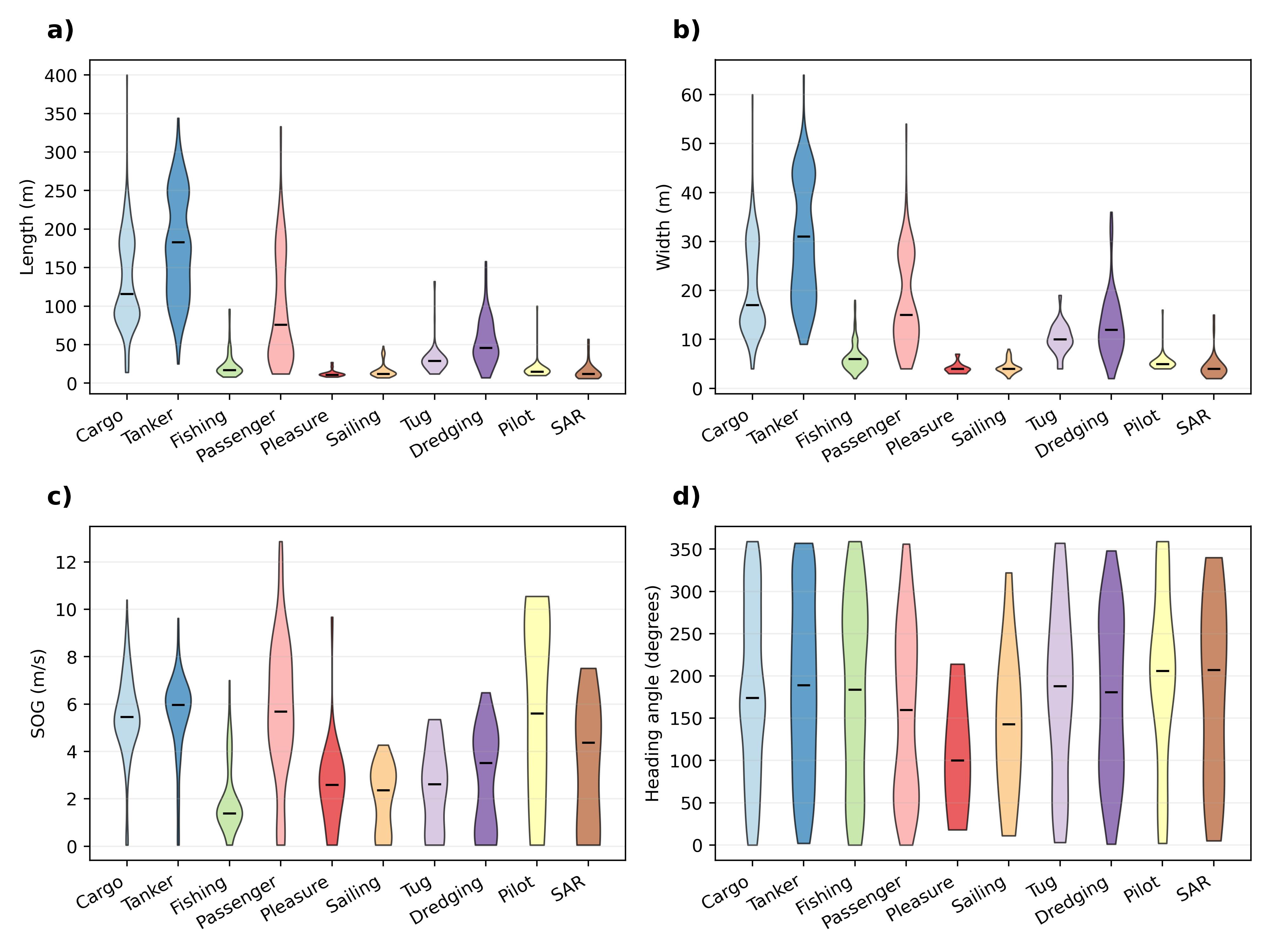} 
    \caption{Statistical distribution of ship characteristics: a) Length, b) Width, c) SOG, d) Heading angle}
    \label{fig:ShapeViolin}
\end{figure}

 Fig.~\ref{fig:ShapeViolin} provides statistical information on ship shapes and movement behavior derived from AIS data, including length, width, SOG, and heading angle. The heading angle represents the direction of the vessel relative to true north, ranging from 0° to 360°. From this figure, it is evident that the dataset contains a wide variety of ship shapes across different types. For instance, there is a significant range and overlap in the length and width of Cargo, Tanker, and Passenger ships. Cargo ships exhibit the broadest range of lengths and widths, whereas Tankers have the highest average length and width among all categories.

\begin{figure}[!t]
    \centering
    \includegraphics[width=0.93\linewidth]{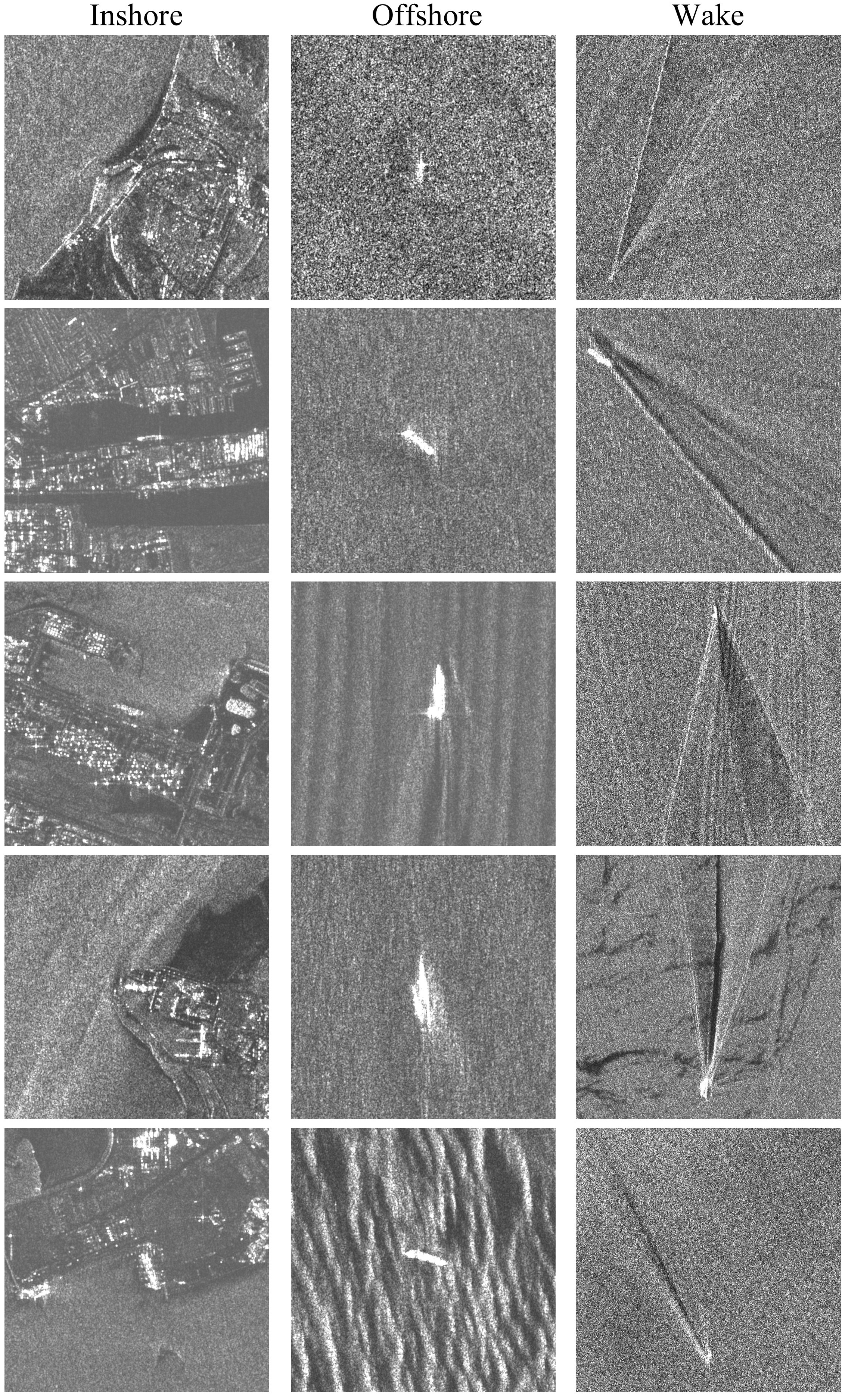}
    \caption{Samples from \datasetname{}: From top to bottom, each row corresponds to fishing boats, cargo, tanker, passenger, and tug, respectively.}
    \label{fig:samples}
\end{figure}

\section{Benchmark Experiments}\label{section:Benchmark Experiments}
To validate and analyze the dataset, we explored several classification scenarios commonly examined in the literature. We focused on four primary ship types: Cargo, Tanker, Fishing, and Passenger. Cargo ships are versatile vessels designed to transport goods and materials across oceans and seas, playing a vital role in global trade. Tankers are specialized ships built to carry bulk liquid cargo, such as crude oil, chemicals, and liquefied natural gas, making them indispensable to the energy and chemical industries. Cargo and tanker ships share several structural and operational similarities, including large hull sizes, similar radar signatures, and standardized navigation patterns. Differentiating between these vessel types is crucial for accurate maritime domain awareness, particularly in regions with high traffic or sensitive geopolitical contexts. However, these similarities pose challenges for distinguishing them using satellite SAR imagery, making this a significant area of ongoing research.

Fishing vessels are used to harvest seafood and are essential to the global food supply chain.
Monitoring and detecting fishing vessels is especially important due to the prevalence of illegal, unreported, and unregulated (IUU) fishing, which threatens marine biodiversity and the sustainability of global fish stocks, while costing legitimate fishermen and governments billions of dollars \cite{paolo2024satellite}. Detecting these vessels is challenging, particularly when they operate in remote areas or disable tracking systems. Their relatively small size and erratic movement patterns further complicate identification using satellite SAR imagery.
Passenger ships, including ferries and cruise liners, are designed to transport people and serve as key components of both public transportation and the tourism sector. Monitoring these ship types is essential for ensuring operational efficiency, environmental compliance, maritime safety, and the protection of marine ecosystems.

We considered five classification scenarios derived from combinations of four primary ship types using benchmark deep learning models, including ResNet, and ResNeXt variants\cite{He2015, Xie2016}, DenseNet121 \cite{huang2017densely}, EfficientNet \cite{tan2021efficientnetv2}, and Vision Transformers \cite{tiny_vit}. 
Each image patch used for modelling is in intensity-format, and is normalised to the $[0,1]$ range and resampled to a $5\times5$\,m effective pixel spacing to reduce computational load, which halves the spatial dimensions of each patch.
The framework, implemented in PyTorch, used Adam optimization with weighted cross-entropy loss to address class imbalance. Training employed a batch size of 16, learning rates between 0.0001–0.0005, and linear decay after five epochs without validation improvement. Ten-fold cross-validation ensured robustness, with 70\% of data for training and validation and 30\% for testing per fold. 

Table~\ref{tab:ResultsInShoreOffshore} presents classification results, comparing model performance when trained on all the ships located in inshore or offshore areas. Inshore environments pose significant challenges due to the presence of land and infrastructure, which introduce additional scattering effects. These areas are characterized by cluttered backgrounds from static objects such as ports and buildings, making it difficult to distinguish ships from surrounding structures. The proximity to land also results in strong backscatter and land-sea boundary confusion, while small vessels with low radar cross-sections are particularly difficult to detect amid high clutter. Furthermore, multipath reflections and shadowing caused by coastal infrastructure can distort radar returns. Offshore areas, by contrast, present different challenges. The low contrast between ships and sea clutter, especially under rough conditions involving strong currents and waves, can produce scattering patterns and speckle noise that mimic or obscure ship signatures, particularly for small or low-profile vessels. Additionally, the motion of ships can introduce distortions such as azimuthal smearing, further complicating detection and classification.

To provide a more detailed analysis, classification results for ships in offshore and inshore areas are presented separately in Table~\ref{tab:ResultsOffShore} and Table~\ref{tab:ResultsInShore}, additionally, per-class analyses are provided in S3 of the Supplementary material. In general, offshore scenarios exhibit better performance, with higher accuracy and precision. On the one hand, this can be attributed to factors as discussed above—namely, the presence of backscattering contributions from land and infrastructure in inshore areas, which can diminish the relative scattering from ships and hinder their identification, especially in lower-resolution imagery. However, this could also be considered indicative of the importance of ship wakes as discriminative features in identifying maritime platforms, since such features are not present inshore. Another contributing factor may also be the smaller number of samples available in inshore regions. This is reflected in the robustness of the offshore results (Table~\ref{tab:ResultsOffShore}), which show lower variation across the ten-fold cross-validation experiments, compared to higher fluctuations in the inshore results (Table~\ref{tab:ResultsInShore}).

Compared with higher-resolution datasets such as OpenSARShip \cite{huang2017opensarship, li2017opensarship2} and FUSAR \cite{hou2020fusar}, the lower performance of NASTaR can be attributed to the smaller dataset size, the moderate resolution of NovaSAR Stripmap imagery, and the mission’s low‑cost design. Nevertheless, the results show clear gains over earlier S-band studies \cite{Kamirul2024, rizaev2022synthwakesar}, which relied heavily on simulations and confirm that S-band measurements provide useful discriminatory information for ship-type classification.


\begin{table}[htbp]
\caption{Classification Results - Inshore and Offshore Scenario}
\centering
\scriptsize
\setlength{\tabcolsep}{2pt}
\begin{tabular}{|l|c|c|c|c|}
\hline
\textbf{Categories} & \textbf{OA} & \textbf{AA} & \textbf{APr} & \textbf{AF1} \\
\hline
Fishing, Other & 87.6 $\pm$ 2.9 & 78.2 $\pm$ 5.5 & 79.0 $\pm$ 4.8 & 77.6 $\pm$ 4.5 \\
Fishing, Cargo & 88.5 $\pm$ 3.3 & 84.7 $\pm$ 4.9 & 87.9 $\pm$ 3.5 & 85.8 $\pm$ 4.4 \\
Cargo, Tanker & 75.4 $\pm$ 1.6 & 65.7 $\pm$ 2.1 & 76.4 $\pm$ 4.0 & 66.7 $\pm$ 2.5 \\
Fishing, Cargo, Tanker & 70.8 $\pm$ 2.4 & 67.8 $\pm$ 1.9 & 71.2 $\pm$ 3.4 & 68.5 $\pm$ 2.4 \\
Fishing, Passenger, Cargo, Tanker & 61.9 $\pm$ 1.5 & 58.1 $\pm$ 2.5 & 64.5 $\pm$ 1.8 & 59.6 $\pm$ 2.1 \\
\hline
\end{tabular}
\label{tab:ResultsInShoreOffshore}
\end{table}

\begin{table}[htbp]
\caption{Classification Results - Offshore Scenario}
\centering
\scriptsize
\setlength{\tabcolsep}{2pt}
\begin{tabular}{|l|c|c|c|c|}
\hline
\textbf{Categories} & \textbf{OA} & \textbf{AA} & \textbf{APr} & \textbf{AF1} \\
\hline
Fishing, Other & 91.1 $\pm$ 2.0 & 76.8 $\pm$ 7.5 & 78.5 $\pm$ 4.8 & 76.5 $\pm$ 5.7 \\
Fishing, Cargo & 91.9 $\pm$ 3.4 & 83.0 $\pm$ 7.2 & 90.0 $\pm$ 4.3 & 85.5 $\pm$ 6.5 \\
Cargo, Tanker & 74.6 $\pm$ 1.8 & 66.4 $\pm$ 2.2 & 76.2 $\pm$ 3.5 & 67.2 $\pm$ 2.6 \\
Fishing, Cargo, Tanker & 70.7 $\pm$ 3.1 & 66.4 $\pm$ 2.0 & 72.2 $\pm$ 4.3 & 68.0 $\pm$ 2.2 \\
Fishing, Passenger, Cargo, Tanker & 61.6 $\pm$ 1.5 & 54.7 $\pm$ 3.6 & 63.1 $\pm$ 2.0 & 56.6 $\pm$ 3.2 \\
\hline
\end{tabular}
\label{tab:ResultsOffShore}
\end{table}

\begin{table}[htbp!]
\caption{Classification Results - Inshore Scenario}
\centering
\scriptsize
\setlength{\tabcolsep}{2pt}
\begin{tabular}{|l|c|c|c|c|}
\hline
\textbf{Categories} & \textbf{OA} & \textbf{AA} & \textbf{APr} & \textbf{AF1} \\
\hline
Fishing, Other & 78.9 $\pm$ 6.8 & 75.9 $\pm$ 4.0 & 77.5 $\pm$ 5.1 & 75.2 $\pm$ 5.6 \\
Fishing, Cargo & 79.2 $\pm$ 5.6 & 78.3 $\pm$ 4.8 & 79.3 $\pm$ 5.4 & 77.7 $\pm$ 5.3 \\
Cargo, Tanker & 75.0 $\pm$ 9.1 & 60.7 $\pm$ 11.6 & 66.5 $\pm$ 17.3 & 60.2 $\pm$ 10.3 \\
Fishing, Cargo, Tanker & 68.5 $\pm$ 2.1 & 57.1 $\pm$ 5.2 & 71.8 $\pm$ 4.7 & 58.4 $\pm$ 5.4 \\
Fishing, Passenger, Cargo, Tanker & 62.6 $\pm$ 1.6 & 53.4 $\pm$ 0.7 & 60.6 $\pm$ 7.5 & 53.9 $\pm$ 2.5 \\
\hline
\end{tabular}
\label{tab:ResultsInShore}
\end{table}

To demonstrate the utility of the wake subset, we performed an additional experiment on the three most challenging classes in NASTaR (Cargo, Passenger, and Tanker). For each vessel with an available wake patch, we trained a 10-fold classifier using (i) the wake patches themselves and (ii) the corresponding main ship patches. Table~\ref{tab:wake_vs_mainship} summarizes the resulting performance.The results show that models trained on wake patches consistently outperform those trained on the corresponding main ship patches across all metrics. This is notable because Cargo, Passenger, and Tanker vessels exhibit substantial overlap in shape and velocity characteristics (see Fig.~\ref{fig:ShapeViolin}), making them difficult to distinguish in moderate-resolution SAR imagery. The wake patterns provide additional discriminative cues, supporting the relevance and usefulness of the dedicated wake subset in NASTaR.

\begin{table}[htbp!]
\centering
\caption{Classification results on Wake dataset} 
\scriptsize
\setlength{\tabcolsep}{4pt}
\begin{tabular}{|l|c|c|c|c|}
\hline
\textbf{Dataset} & \textbf{OA} & \textbf{AA} & \textbf{APr} & \textbf{AF1} \\
\hline
Ship+Wake 
 & $57.71 \pm 2.01$ 
 & $48.05 \pm 2.71$ 
 & $58.68 \pm 4.74$ 
 & $48.78 \pm 3.63$ \\
\hline
Ship-only 
 & $52.38 \pm 4.11$ 
 & $42.90 \pm 3.47$ 
 & $48.79 \pm 5.96$ 
 & $41.94 \pm 4.16$ \\
\hline
\end{tabular}
\label{tab:wake_vs_mainship}
\end{table}

\section{Conclusion}\label{section:Conclusion}
This paper introduces \datasetname{}, a dedicated dataset for ship-type classification, created by leveraging NovaSAR S-band imagery and AIS-derived labels. To build this dataset, we designed a detailed workflow to meticulously extract high-quality samples from SAR scenes and their corresponding AIS data. We also conducted comprehensive statistical analyses to demonstrate the dataset’s versatility.
To facilitate future research, \datasetname{} includes additional features such as ship-to-shore distance, inshore/offshore categorization, and a separate wake dataset for patches where ship wakes are visible, enabling further wake-related studies. The dataset’s applicability was validated across prominent ship-type classification scenarios using benchmark deep learning models. Results show promising performance: over 60\% accuracy for four major ship types, over 70\% for a three-class scenario, more than 75\% for distinguishing cargo from tanker ships, and over 87\% for identifying fishing vessels.
Potential future research includes exploring cross-domain and multi-platform learning, such as assessing the transferability of features across different SAR frequencies and platforms, designing tailored deep learning architectures to improve benchmark results or further investigation of the impact of wake patterns on sea surface characteristics and their role in maritime activity monitoring.
\section*{Acknowledgment}

The authors would like to thank Surrey Satellite Technology Ltd. (SSTL) and Airbus UK for the kind provision of NovaSAR image products. Copyright $\copyright$ 2025 SSTL.

\ifCLASSOPTIONcaptionsoff
  \newpage
\fi



\bibliographystyle{IEEEtran}
\bibliography{bibtex/bib/References}

@article{xing2013ship,
  title={Ship classification in {T}erra{SAR}-{X} images with feature space based sparse representation},
  author={Xing, Xiangwei and Ji, Kefeng and Zou, Huanxin and Chen, Wenting and Sun, Jixiang},
  journal={IEEE Geosci. Remote Sens. Lett.},
  volume={10},
  number={6},
  pages={1562--1566},
  year={2013},
  publisher={IEEE}
}

@article{huang2017opensarship,
  title={{Open{SAR}Ship: A dataset dedicated to {S}entinel-1 ship interpretation}},
  author={Huang, Lanqing and Liu, Bin and Li, Boying and Guo, Weiwei and Yu, Wenhao and Zhang, Zenghui and Yu, Wenxian},
  journal={IEEE J. Sel. Topics Appl. Earth Observ. Remote Sens.},
  volume={11},
  number={1},
  pages={195--208},
  year={2017},
  publisher={IEEE}
}

@inproceedings{li2017opensarship2,
  title={Open{SAR}{S}hip 2.0: {A} large-volume dataset for deeper interpretation of ship targets in {S}entinel-1 imagery},
  author={Li, Boying and Liu, Bin and Huang, Lanqing and Guo, Weiwei and Zhang, Zenghui and Yu, Wenxian},
  booktitle={2017 {SAR} in Big Data Era: Models, Methods and Applications (BIGSARDATA)},
  pages={1--5},
  year={2017},
  organization={IEEE}
}

@INPROCEEDINGS{Kamirul2024,
  author={Kamirul, Kamirul and Pappas, Odysseas and Rizaev, Igor G. and Achim, Alin},
  booktitle={2024 IEEE Int. Geosci. Remote Sens. Symp. (IGARSS)}, 
  title={On the modelling of ship wakes in {S}-Band {SAR} Images and an Application to Ship Identification}, 
  year={2024},
  volume={},
  number={},
  pages={10599-10603},
  doi={10.1109/IGARSS53475.2024.10642130}
}

@article{hou2020fusar,
  title={{FUSAR}-{S}hip: Building a high-resolution {SAR}-{AIS} matchup dataset of {G}aofen-3 for ship detection and recognition},
  author={Hou, Xiyue and Ao, Wei and Song, Qian and Lai, Jian and Wang, Haipeng and Xu, Feng},
  journal={Sci. China Inf. Sci.},
  volume={63},
  number={4},
  pages={140303},
  year={2020},
  publisher={Springer}
}

@article{ma2018ship,
  title={Ship classification and detection based on {CNN} using {GF}-3 {SAR} images},
  author={Ma, Mengyuan and Chen, Jie and Liu, Wei and Yang, Wei},
  journal={Remote Sensing},
  volume={10},
  number={12},
  pages={2043},
  year={2018},
  publisher={MDPI}
}

@article{xu2024opensarwake,
  title={Open{SAR}{W}ake: A large-scale {SAR} dataset for ship wake recognition with a feature refinement oriented detector},
  author={Xu, Chengji and Wang, Xiaoqing},
  journal={IEEE Geosci. Remote Sens. Lett.},
  volume={21},
  pages={1--5},
  year={2024},
  publisher={IEEE}
}

@article{rizaev2022synthwakesar,
  title={SynthWake{SAR}: A synthetic {SAR} dataset for deep learning classification of ships at sea},
  author={Rizaev, Igor G and Achim, Alin},
  journal={Remote Sensing},
  volume={14},
  number={16},
  pages={3999},
  year={2022},
  publisher={MDPI}
}

@article{zhang2021imbalanced,
  title={Imbalanced high-resolution {SAR} ship recognition method based on a lightweight {CNN}},
  author={Zhang, Ying and Lei, Zhiyong and Yu, Hui and Zhuang, Long},
  journal={IEEE Geosci. Remote Sens. Lett.},
  volume={19},
  pages={1--5},
  year={2021},
  publisher={IEEE}
}

@article{wang2023sar,
  title={{SAR} ship target recognition via multiscale feature attention and adaptive-weighed classifier},
  author={Wang, Chenwei and Pei, Jifang and Luo, Siyi and Huo, Weibo and Huang, Yulin and Zhang, Yin and Yang, Jianyu},
  journal={IEEE Geosci. Remote Sens. Lett.},
  volume={20},
  pages={1--5},
  year={2023},
  publisher={IEEE}
}

@article{zheng2022metaboost,
  title={Meta{B}oost: A novel heterogeneous {DCNN}s ensemble network with two-stage filtration for {SAR} ship classification},
  author={Zheng, Hao and Hu, Zhigang and Liu, Jianjun and Huang, Yuhang and Zheng, Meiguang},
  journal={IEEE Geosci. Remote Sens. Lett.},
  volume={19},
  pages={1--5},
  year={2022},
  publisher={IEEE}
}

@misc{DMA_AIS_Data,
  author = {{Danish Maritime Authority}},
  title = {{Historical AIS Data}},
  note = {Accessed: 25 Nov. 2025},
  howpublished = {Available online at \url{http://aisdata.ais.dk/} }
}

@misc{Natural_Earth_vec,
  author = {{Natural Earth}},
  title = {Natural Earth Vector 10m map},
  note = {Accessed: 25 Nov. 2025},
  howpublished = {Available online at \url{https://www.naturalearthdata.com/} }
}

@article{paolo2024satellite,
  title={Satellite mapping reveals extensive industrial activity at sea},
  author={Paolo, Fernando S and Kroodsma, David and Raynor, Jennifer and Hochberg, Tim and Davis, Pete and Cleary, Jesse and Marsaglia, Luca and Orofino, Sara and Thomas, Christian and Halpin, Patrick},
  journal={Nature},
  volume={625},
  number={7993},
  pages={85--91},
  year={2024},
  publisher={Nature Publishing Group UK London}
}

@article{zhang2024development,
  title={Development and application of ship detection and classification datasets: A review},
  author={Zhang, Chi and Zhang, Xi and Gao, Gui and Lang, Haitao and Liu, Genwang and Cao, Chenghui and Song, Yuying and Guan, Yanan and Dai, Yongshou},
  journal={IEEE Geosci. Remote Sens. Mag.},
  year={2024},
  publisher={IEEE}
}

@article{awais2025survey,
  title={A Survey on {SAR} ship classification using Deep Learning},
  author={Awais, Ch Muhammad and Reggiannini, Marco and Moroni, Davide and Salerno, Emanuele},
  journal={arXiv preprint arXiv:2503.11906},
  year={2025}
}

@article{paolo2022xview3,
  title={{xView3-SAR}: Detecting dark fishing activity using {S}ynthetic {A}perture {R}adar imagery},
  author={Paolo, Fernando and Lin, Tsu-ting Tim and Gupta, Ritwik and Goodman, Bryce and Patel, Nirav and Kuster, Daniel and Kroodsma, David and Dunnmon, Jared},
  journal={Adv. Neural Inf. Process. Syst.},
  volume={35},
  pages={37604--37616},
  year={2022}
}

@inproceedings{huang2017densely,
  title={Densely Connected Convolutional Networks},
  author={Huang, Gao and Liu, Zhuang and van der Maaten, Laurens and Weinberger, Kilian Q },
  booktitle={Proceedings of the IEEE Conference on Computer Vision and Pattern Recognition},
  year={2017}
}

@article{He2015,
  author = {Kaiming He and Xiangyu Zhang and Shaoqing Ren and Jian Sun},
  title = {Deep Residual Learning for Image Recognition},
  journal = {arXiv preprint arXiv:1512.03385},
  year = {2015}
}

@article{Xie2016,
  title={Aggregated Residual Transformations for Deep Neural Networks},
  author={Saining Xie and Ross Girshick and Piotr Dollár and Zhuowen Tu and Kaiming He},
  journal={arXiv preprint arXiv:1611.05431},
  year={2016}
}

@inproceedings{tan2021efficientnetv2,
  title={Efficientnetv2: Smaller models and faster training},
  author={Tan, Mingxing and Le, Quoc},
  booktitle={International conference on machine learning},
  pages={10096--10106},
  year={2021},
  organization={PMLR}
}

@InProceedings{tiny_vit,
  title={{T}iny{V}i{T}: Fast Pretraining Distillation for Small Vision Transformers},
  author={Wu, Kan and Zhang, Jinnian and Peng, Houwen and Liu, Mengchen and Xiao, Bin and Fu, Jianlong and Yuan, Lu},
  booktitle={European conference on computer vision (ECCV)},
  year={2022}
}
\end{document}


\begin{center}
    {\Large \textbf{Supplementary Material}}\\[6pt]
    {\large for the manuscript}\\[4pt]
    {\large \textit{“NASTaR: A NovaSAR-Based Automated Ship Target Recognition Dataset”}}\\[12pt]
    \rule{0.9\linewidth}{0.4pt}
\end{center}

\section*{S1. SAR--AIS Matching and Filtering Algorithm}



\begin{algorithm}[H]
\caption{SAR--AIS Matching and Filtering Pipeline}
\label{alg:sar_ais_matching}
\begin{algorithmic}[1]

\State \textbf{Input:} SAR image $I$ with acquisition time $t_{\text{SAR}}$ and footprint region $\Omega$.
\State \textbf{Output:} Cleaned set of ship patches with corresponding AIS metadata.

\Statex
\State \textbf{/* AIS Extraction */}
\State Extract all AIS messages whose latitude/longitude fall inside $\Omega$ and whose 
timestamps lie within $t_{\text{SAR}} \pm 100\ \text{seconds}$.
\State For each unique vessel (identified by name or MMSI), keep only the AIS record
closest in time to $t_{\text{SAR}}$.

\Statex
\State \textbf{/* Patch Extraction */}
\For{each remaining AIS record}
    \State Use the reported $(\text{lat}, \text{lon})$ to locate the ship on the SAR image.
    \State Crop a fixed-size patch (e.g., $512 \times 512$ pixels) centred at the ship location.
    \State Store the patch and its AIS metadata in the dataset.
\EndFor

\Statex
\State \textbf{/* Duplicate Removal */}
\For{each spatial region where multiple patches overlap}
    \If{all detected ships belong to the same class}
        \State Keep only one representative patch; discard the rest.
    \Else
        \State Discard all patches in that region to avoid ambiguous labelling.
    \EndIf
\EndFor

\Statex
\State \textbf{/* Low-quality Filtering */}
\For{each extracted patch}
    \If{the ship signature is not visible or the image is severely noisy/distorted}
        \State Remove the patch from the dataset.
    \EndIf
\EndFor

\Statex
\State \Return Cleaned set of patches and associated AIS metadata.

\end{algorithmic}
\end{algorithm}

\section*{S2. Additional Dataset Figures}


\begin{figure}[H]
    \centering

    \begin{subfigure}{0.48\linewidth}
        \centering
        \includegraphics[width=\linewidth]{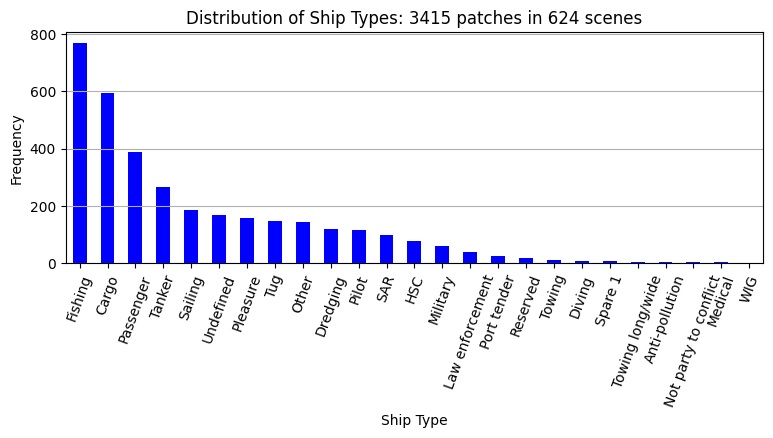}
        \caption{Full dataset (before filtering)}
        \label{fig:class_dist_full}
    \end{subfigure}
    \hfill
    \begin{subfigure}{0.48\linewidth}
        \centering
        \includegraphics[width=\linewidth]{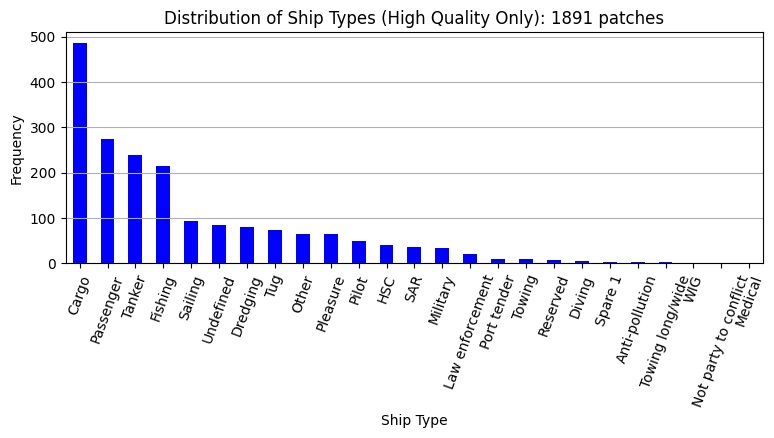}
        \caption{After high-quality filtering}
        \label{fig:class_dist_after}
    \end{subfigure}

    \vspace{0.6cm}

    \begin{subfigure}{0.48\linewidth}
        \centering
        \includegraphics[width=\linewidth]{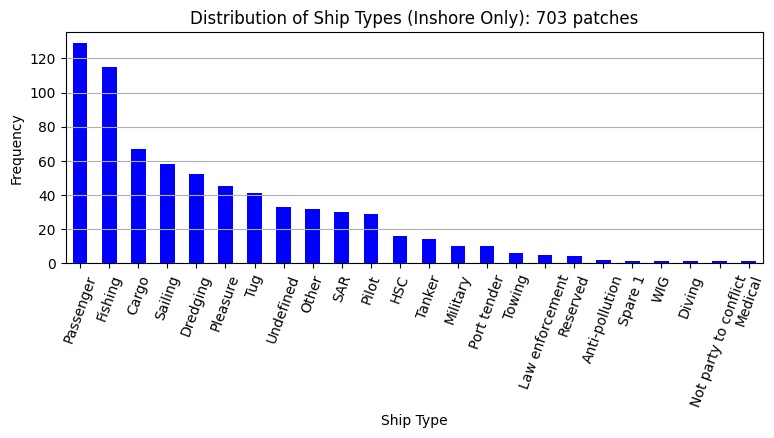}
        \caption{High-quality inshore subset}
        \label{fig:class_dist_inshore}
    \end{subfigure}
    \hfill
    \begin{subfigure}{0.48\linewidth}
        \centering
        \includegraphics[width=\linewidth]{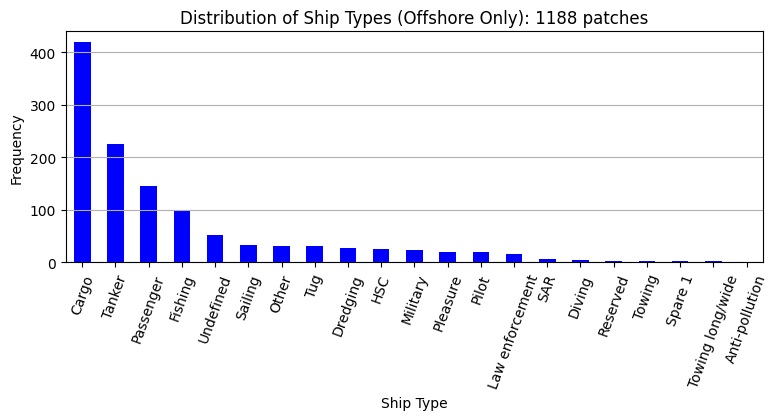}
        \caption{High-quality offshore subset}
        \label{fig:class_dist_offshore}
    \end{subfigure}

    \caption{Class counts of the NASTaR dataset at different stages and subsets.}
    \label{fig:class_distribution_all}
\end{figure}

\begin{figure}[H]
    \centering
    \includegraphics[width=0.85\linewidth]{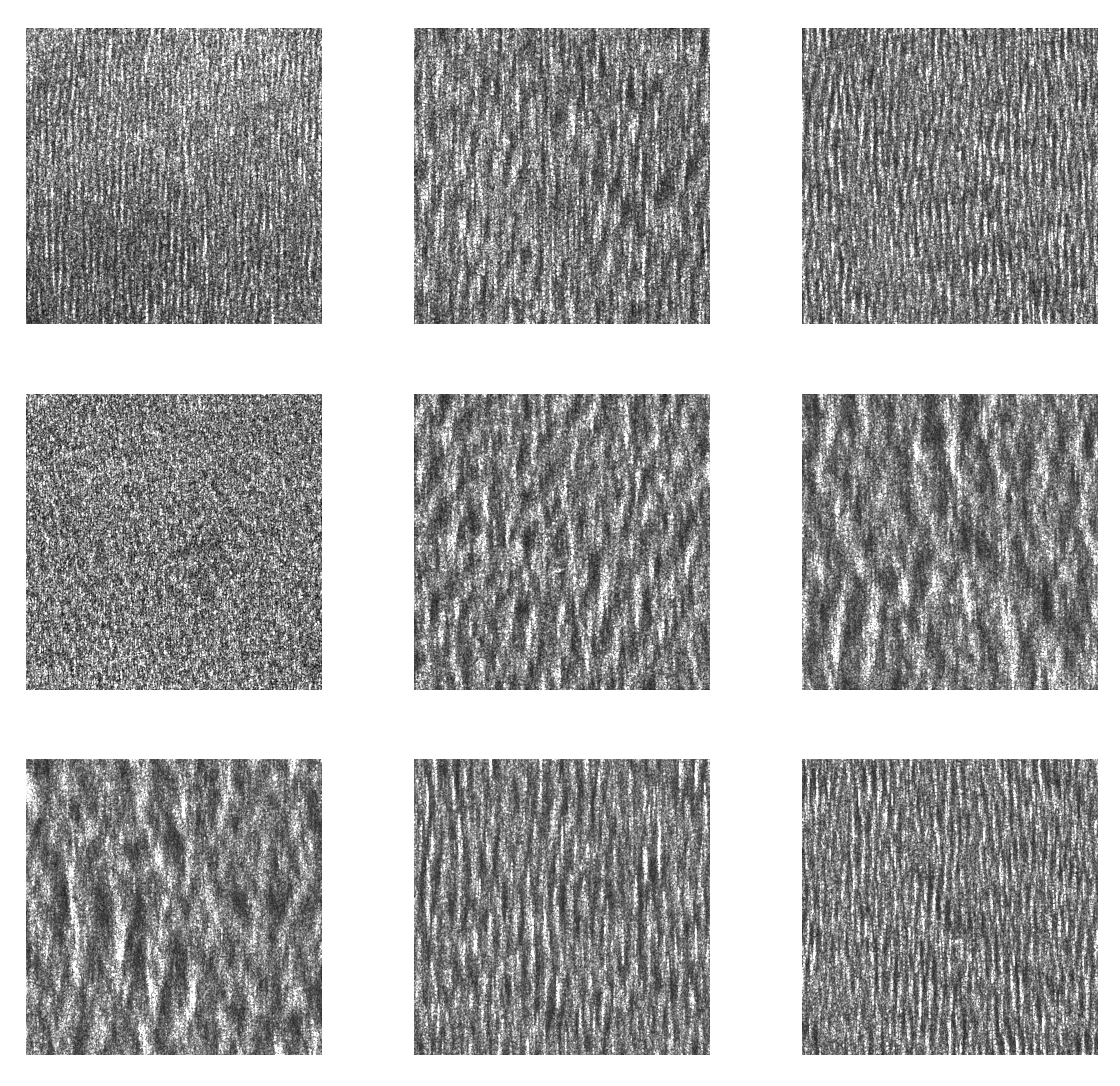}
    \caption{Examples of rejected ship patches during the quality filtering stage.}
    \label{fig:wake_examples}
\end{figure}

\begin{table}[H]
\centering
\caption{Description of dataset fields included in NASTaR.}
\begin{tabular}{p{3.5cm} p{10cm}}
\toprule
\textbf{Field Name} & \textbf{Description} \\
\midrule

Timestamp & Time at which the AIS message was recorded. \\

Type of mobile & AIS transmitter category (e.g., Class A, Class B, Base Station, AtoN). \\

MMSI & Maritime Mobile Service Identity; unique vessel identifier used for matching AIS to SAR patches. \\

Latitude, Longitude & Geolocation of the vessel at the AIS timestamp. Used to locate the patch in the SAR image. \\

Navigational status & Vessel activity (e.g., Under way using engine, Moored, Restricted manoeuvrability). \\

ROT & Rate of turn reported by AIS (degrees per minute). \\

SOG & Speed Over Ground (knots); used to identify potential wake-generating ships. \\

COG & Course Over Ground (degrees); represents the vessel’s direction of movement. \\

Heading & Ship’s orientation relative to true north. \\

IMO & International Maritime Organization identifier (if available). \\

Callsign & Vessel radio callsign (if available). \\

Name & Ship name provided in AIS messages. \\

Ship type & AIS-based vessel class (Cargo, Tanker, Fishing, Passenger, etc.). Used as the primary classification label. \\

Cargo type & Additional AIS cargo-category information (if present). \\

Width, Length & Physical dimensions of the vessel (meters), used for dataset statistics. \\

Type of position fixing device & AIS-reported navigation system used (e.g., GPS). \\

Draught & Vessel draught (meters), when available. \\

Destination & Vessel’s declared destination from AIS. \\

ETA & Estimated time of arrival at destination. \\

Data source type & Indicates whether AIS data was received via terrestrial or satellite receivers. \\

A, B, C, D & Standard AIS message geometry parameters (reference points for ship shape). \\

Patch\_name & Unique filename of the extracted SAR patch associated with the AIS record. \\

Dist\_to\_land & Distance (meters) from vessel to nearest shoreline, used to determine inshore/offshore categorization. \\

Shoreline & Inshore/Offshore label derived from distance-to-land threshold. \\

Quality & Filtering label indicating whether the patch passed the quality check (1 = accepted, 0 = removed). \\

Wake Potential & Binary indicator derived from SOG, denoting likelihood of a visible wake (0 and 1). \\

Wake Quality & Score indicating clarity/strength of the visible wake pattern, if present (0 and 1). \\

\bottomrule
\end{tabular}
\label{tab:dataset_fields}
\end{table}

\section*{S3. Confusion Matrices and Per-Class Results}



\begin{table}[H]
\centering
\caption{Aggregated confusion matrix from all experiments with precision, recall, and F1-score for the 2-class (Fishing vs Other) scenario.}
\begin{tabular}{c|cc|c}
\toprule
 & \multicolumn{2}{c|}{\textbf{Predicted Class}} & \\
\textbf{True Class} & Fishing & Other & \textbf{Recall} \\
\midrule
\textbf{Precision} 
& $0.6497 \pm 0.1054$ 
& $0.9301 \pm 0.0221$ 
& \\ 
\midrule
Fishing & 347 & 193 & $0.6426 \pm 0.1383$ \\
Other   & 214 & 2516 & $0.9216 \pm 0.0476$ \\
\midrule
\textbf{F1} 
& $0.6268 \pm 0.0782$ 
& $0.9247 \pm 0.0193$ 
& \\
\bottomrule
\end{tabular}
\end{table}

\begin{table}[H]
\centering
\caption{Aggregated confusion matrix from all experiments with precision, recall, and F1-score for the 2-class (Cargo vs Fishing) scenario.}
\begin{tabular}{c|cc|c}
\toprule
 & \multicolumn{2}{c|}{\textbf{Predicted Class}} & \\
\textbf{True Class} & Cargo & Fishing & \textbf{Recall} \\
\midrule
\textbf{Precision} 
& $0.8967 \pm 0.0356$ 
& $0.8606 \pm 0.0578$ 
& \\ 
\midrule
Cargo   & 1152 & 68  & $0.9443 \pm 0.0302$ \\
Fishing & 135  & 405 & $0.7500 \pm 0.1002$ \\
\midrule
\textbf{F1} 
& $0.9192 \pm 0.0224$ 
& $0.7967 \pm 0.0668$ 
& \\
\bottomrule
\end{tabular}
\end{table}

\begin{table}[H]
\centering
\caption{Aggregated confusion matrix from all experiments with precision, recall, and F1-score for the 2-class (Cargo vs Tanker) scenario.}
\begin{tabular}{c|cc|c}
\toprule
 & \multicolumn{2}{c|}{\textbf{Predicted Class}} & \\
\textbf{True Class} & Cargo & Tanker & \textbf{Recall} \\
\midrule
\textbf{Precision} 
& $0.7531 \pm 0.0130$ 
& $0.7747 \pm 0.0792$ 
& \\ 
\midrule
Cargo  & 1151 & 69  & $0.9434 \pm 0.0270$ \\
Tanker & 378  & 222 & $0.3700 \pm 0.0515$ \\
\midrule
\textbf{F1} 
& $0.8373 \pm 0.0117$ 
& $0.4964 \pm 0.0444$ 
& \\
\bottomrule
\end{tabular}
\end{table}

\begin{table}[H]
\centering
\caption{Aggregated confusion matrix from all experiments with precision, recall, and F1-score for the 3-class (Cargo, Fishing, Tanker) scenario.}
\begin{tabular}{c|ccc|c}
\toprule
 & \multicolumn{3}{c|}{\textbf{Predicted Class}} & \\
\textbf{True Class} & Cargo & Fishing & Tanker & \textbf{Recall} \\
\midrule
\textbf{Precision} 
& $0.6979 \pm 0.0157$ 
& $0.7629 \pm 0.0485$ 
& $0.6745 \pm 0.0633$ 
& \\ 
\midrule
Cargo   & 978 & 116 & 126 & $0.8016 \pm 0.0536$ \\
Fishing & 111 & 423 & 6   & $0.7833 \pm 0.0673$ \\
Tanker  & 312 & 18  & 270 & $0.4500 \pm 0.0522$ \\
\midrule
\textbf{F1} 
& $0.7455 \pm 0.0289$ 
& $0.7705 \pm 0.0383$ 
& $0.5382 \pm 0.0510$ 
& \\
\bottomrule
\end{tabular}
\end{table}

\begin{table}[H]
\centering
\caption{Aggregated confusion matrix from all experiments with precision, recall, and F1-score for the 4-class (Cargo, Fishing, Passenger, Tanker) scenario.}
\begin{tabular}{c|cccc|c}
\toprule
 & \multicolumn{4}{c|}{\textbf{Predicted Class}} & \\
\textbf{True Class}
 & Cargo & Fishing & Passenger & Tanker & \textbf{Recall} \\
\midrule
\textbf{Precision}
 & $0.5801 \pm 0.0200$
 & $0.6997 \pm 0.0216$
 & $0.6491 \pm 0.0485$
 & $0.6491 \pm 0.0853$
 & \\
\midrule
Cargo     & 952 & 58  & 105 & 105 & $0.7803 \pm 0.0541$ \\
Fishing   & 122 & 345 & 66  & 7   & $0.6389 \pm 0.1160$ \\
Passenger & 242 & 77  & 353 & 18  & $0.5116 \pm 0.0794$ \\
Tanker    & 328 & 13  & 22  & 237 & $0.3950 \pm 0.0435$ \\
\midrule
\textbf{F1}
 & $0.6643 \pm 0.0221$
 & $0.6624 \pm 0.0681$
 & $0.5683 \pm 0.0556$
 & $0.4908 \pm 0.0574$
 & \\
\bottomrule
\end{tabular}
\label{tab:confmat_4class}
\end{table}

\subsection*{S3.1 Per-class results on Wake dataset experiments}

\begin{table}[H]
\centering
\caption{Aggregated confusion matrix from all experiments with precision, recall, and F1-score for the 3-class (Cargo, Passenger, Tanker) scenario from the Wake dataset.}
\begin{tabular}{c|ccc|c}
\toprule
 & \multicolumn{3}{c|}{\textbf{Predicted Class}} & \\
\textbf{True Class} 
& Cargo & Passenger & Tanker & \textbf{Recall} \\
\midrule
\textbf{Precision} 
& $0.5734 \pm 0.0108$ 
& $0.6063 \pm 0.1077$ 
& $0.5808 \pm 0.1053$ 
& \\ 
\midrule
Cargo     
& 527 & 35 & 58 
& $0.8500 \pm 0.0468$ \\
Passenger 
& 157 & 75 & 18 
& $0.3000 \pm 0.1032$ \\
Tanker    
& 235 & 13 & 102 
& $0.2914 \pm 0.0475$ \\
\midrule
\textbf{F1}
& $0.6844 \pm 0.0196$ 
& $0.3936 \pm 0.1102$
& $0.3856 \pm 0.0548$ 
& \\
\bottomrule
\end{tabular}
\label{tab:confmat_3class}
\end{table}

\begin{table}[H]
\centering
\caption{Aggregated confusion matrix from all experiments, together with precision, recall, and F1‑score, for the 3‑class (Cargo, Passenger, Tanker) scenario corresponding to Wake‑dataset ships, evaluated using their main ship patches.}
\begin{tabular}{c|ccc|c}
\toprule
 & \multicolumn{3}{c|}{\textbf{Predicted Class}} & \\
\textbf{True Class}
 & Cargo & Passenger & Tanker & \textbf{Recall} \\
\midrule
\textbf{Precision}
 & $0.5625 \pm 0.0276$
 & $0.4441 \pm 0.1299$
 & $0.4570 \pm 0.0949$
 & \\
\midrule
Cargo     
 & 490 & 57 & 73 
 & $0.7903 \pm 0.1235$ \\
Passenger 
 & 151 & 62 & 37 
 & $0.2480 \pm 0.1250$ \\
Tanker    
 & 230 & 33 & 87 
 & $0.2486 \pm 0.0746$ \\
\midrule
\textbf{F1}
 & $0.6532 \pm 0.0580$
 & $0.2913 \pm 0.1023$
 & $0.3137 \pm 0.0601$
 & \\
\bottomrule
\end{tabular}
\label{tab:confmat_newfile_3class}
\end{table}

\newpage 

\section*{S4. AIS--SAR Association Uncertainty Analysis}

The reliability of AIS--SAR matching depends primarily on (i) the latency of AIS
reporting, (ii) vessel displacement between AIS transmission and SAR image 
acquisition, (iii) AIS GPS positional error, and (iv) NovaSAR geolocation 
uncertainty.

\subsection*{S4.1 AIS Reporting Latency}
The Danish Maritime Authority (DMA) AIS feed provides position reports at a 
10\,s interval (as observed in our AIS timestamps). Let the ship's speed be 
$v$ (in knots). A vessel travelling at speed $v$ moves approximately
\[
d_{\mathrm{AIS}} = 0.514 \, v \times 10 \;\text{m}.
\]

For a typical vessel in our dataset (15\,knots):
\[
d_{\mathrm{AIS}} 
= 0.514 \times 15 \times 10 
\approx 77 \;\text{m}.
\]

For a fast vessel (30\,knots):
\[
d_{\mathrm{AIS}} 
= 0.514 \times 30 \times 10 
\approx 154 \;\text{m}.
\]

\subsection*{S4.2 AIS GPS Positional Uncertainty}
AIS positional information originates from the vessel's onboard GPS. 
Typical civilian GPS horizontal accuracy is within $\approx 10$\,m. 
We therefore denote:
\[
\epsilon_{\mathrm{GPS}} \approx 10\;\text{m}.
\]

\subsection*{S4.3 NovaSAR Geolocation Uncertainty}
NovaSAR Stripmap products have a ground sampling distance of 6\,m, and we have respaced out products to 2.5\,m. Product geolocation is performed via on-board GPS or TLE orbits. Mission requirement for geolocation errors is under 50\,m, with mean geolocation errors reported  as typically below 10\,m in the official mission documentation\footnote{CSIRO, ``NovaSAR-1 User Guide.'' \url{https://research.csiro.au/cceo/novasar/about/novasar-1-user-guide}, (accessed March $17^{th}$, 2026)}.


Additionally, to further refine the geolocation accuracy in our dataset, we manually extracted more than twenty tie-points per scene and matched them using 2D Affine transformation and are confident this brings any remaining geolocation inaccuracy below 5\,m. Thus,
\[
\epsilon_{\mathrm{SAR}} < 5\;\text{m}.
\]

\subsection*{S4.4 Total AIS--SAR Association Error}
Assuming independent uncertainties, the combined spatial uncertainty is
\[
\epsilon_{\mathrm{total}} 
= \sqrt{ d_{\mathrm{AIS}}^2 
        + (\epsilon_{\mathrm{GPS}} + \epsilon_{\mathrm{SAR}})^2 }.
\]

For a 15\,kn vessel:
\[
\epsilon_{\mathrm{total}} 
\approx \sqrt{ 77^2 + 15^2 } 
\approx 78.4 \;\text{m}.
\]

For a 30\,kn vessel:
\[
\epsilon_{\mathrm{total}}
\approx \sqrt{ 154^2 + 15^2 }
\approx 154.7 \;\text{m}.
\]

\subsection*{S4.5 Practical Impact on Patch Extraction}
Each NASTaR patch covers $1280 \times 1280$\,m$^2$ 
(derived from a $512 \times 512$ pixel window at 2.5\,m/pixel). 
Even in the worst case ($\approx 155$\,m), the vessel remains well inside the
patch region. Thus, AIS--SAR uncertainty does not compromise the ship's 
inclusion within the image patch.

Nevertheless, ambiguous cases may arise when multiple ships appear close 
together. As described in Algorithm~S1, we remove all patches containing 
multiple closely spaced ships to avoid potential AIS--SAR misassociation. 
Additionally, manual inspection is performed to remove any patches where the 
ship signature is not visible or is severely degraded.